\documentclass[12pt]{ri_thesis}

\usepackage{amssymb,bm} 

\usepackage{microtype}
\usepackage{graphicx}
\usepackage{epigraph}
\usepackage[cmex10]{amsmath}
\usepackage{amsfonts}
\usepackage{appendix}
\usepackage[numbers,sort]{natbib}
\usepackage{setspace}
\usepackage{layout}
\usepackage[subfigure]{tocloft}  

\usepackage{pgfplots}
\usepackage{pgfplotstable}
\usepackage{tikz}
\pgfplotsset{compat=1.9}
\usetikzlibrary{positioning}
\usetikzlibrary{backgrounds}
\usetikzlibrary{calc}
\usepgfplotslibrary{colorbrewer}  
\usepgfplotslibrary{statistics}

\usepackage[format=hang]{subfig}  
\usepackage{xcolor}  
\usepackage{algorithm}
\usepackage{algpseudocode} 
\usepackage{multirow}

\usepackage{xspace} 
\usepackage{siunitx}
\usepackage[raggedright]{titlesec}  

\usepackage{float}
\newfloat{algorithm}{tbp}{lop}

\colorlet{documentLinkColor}{blue}
\colorlet{documentCitationColor}{black!80}
\usepackage[backref,
        pageanchor=true,
        plainpages=false,
        pdfpagelabels,
        bookmarks,
        bookmarksnumbered,
]{hyperref}
\hypersetup{
     colorlinks = true,
     citecolor = documentCitationColor,
     linkcolor = documentLinkColor,
     urlcolor = documentLinkColor,
}

\usepackage[nameinlink]{cleveref}

\usepackage{fancyhdr}

\definecolor{headergray}{rgb}{0.5,0.5,0.5}

\fancypagestyle{plain}{%
  \fancyhf{}
  \fancyfoot[RO,LE]{\thepage}
}

\fancypagestyle{thesis}{%
  \fancyhead{}%
  \fancyhead[RO,LE]{\hyperlink{contents}{\slshape \leftmark}}
  \fancyfoot[RO,LE]{\thepage}%
}

\newcommand{\chapternote}[1]{{%
  \let\thempfn\relax
  \footnotetext[0]{\emph{#1}}
}}

\usepackage{cleveref}
\usepackage{stackengine}

\newcommand{\figref}[1]{Figure \ref{#1}}
\newcommand{\algoref}[1]{Algorithm \ref{#1}}

\newcommand{\sbf}{\mathbf{s}} 

\newcommand{\Pb}{\mathbf{P}}
\newcommand{\lmdb}{\bm\lambda}
\newcommand{\p}{\text{p}}

\usepackage[
  reset,
  letterpaper,
  twoside,
  vscale=.75, 
  hscale=.70,  
  nomarginpar,  
  heightrounded,  
]{geometry}

\begin {document}

\frontmatter
\pagestyle{empty}

\title{\textbf{Predicting Human Trajectories by Learning and Matching Patterns}\\ \vspace{12pt}}
\author{Dapeng Zhao}
\date{April 04, 2021}
\Year{2021}
\trnumber{CMU-RI-TR-21-07}

\committee{
Dr. Jean Oh, \emph{chair} \\
Dr. John M. Dolan \\
Dr. Heather Jones \\
Jay Patrikar
}

\support{}
\disclaimer{}

\permission{All rights reserved.}

\maketitle


\pagestyle{plain} 

\begin{abstract} 

\setlength{\parskip}{1em}
\setlength{\parindent}{0em}

\noindent
As more and more robots are envisioned to cooperate with humans sharing the same space, it is desired for robots to be able to predict others' trajectories to navigate in a safe and self-explanatory way. 

We propose a Convolutional Neural Network-based approach to learn, detect, and extract patterns in sequential trajectory data, known here as Social Pattern Extraction Convolution (Social-PEC). 

A set of experiments carried out on the human trajectory prediction problem shows that our model performs comparably to the state of the art and outperforms in some cases. More importantly, the proposed approach unveils the obscurity in the previous use of a pooling layer, presenting a way to intuitively explain the decision-making process.  \end{abstract}
\begin{acknowledgments} 

\setlength{\parskip}{1em}
\setlength{\parindent}{0em}

\noindent
There are so many people that I need to forever be grateful for in this journey. 

First, my supervisor, Dr. Jean Oh. She believed in my potentials when I had little to no experience in this research area, and offered me the opportunity to work with her. In the past two years, she almost gave the perfect amount of flexibility and guidance for me to explore and to focus. Beyond my research work, she also genuinely cares about my personal well-being and future career goals. She always asks us to have enough rest and take breaks and vacations, she encouraged me to take internship and pursue career directions that I am passionate about even when they are not aligned with the lab's interests, she sacrificed her personal time to meet with me daily in the beginning of the quarantine to motivate and encourage me when I needed it the most... I will never forget all of Jean's kindness and support to me. 

Second, Dr. Red Whittaker for offering me the full-time position where I started my journey in CMU; Dr. John M. Dolan, Dr. Kris Kitani and Dr. Chieko Asakawa for trusting me and hiring me when my positions keep being terminated for various uncontrollable factors. 

Third, Dr. Ji Zhang for helping me with robot building with his expertise; my lab mates Sam Shum, Advaith Sethuraman, Felix Labelle, Brendon Bolt, Xinjie Yao for their assistance and inspirations to my research. 

I also would like to thank my parents for all of their sacrifice, investment, expectation, support and love for me; my dearest roommates/landlords/best friends in Pittsburgh, David and Allison Lambacher, for taking care of me like a family whenever I am in need; friends from my small group and my church North Way Christian Community, for all the good time and spiritual support and accountability that I needed. 

Lastly, I have to give all credits to God, who created me, forgave all my sins, and loves me unconditionally. If I had any talents or wisdom that made this work possible, they all come from the creator. 
 \end{acknowledgments}
\begin{funding} 

\setlength{\parskip}{1em}
\setlength{\parindent}{0em}

This work is in part supported by the U.S. Air Force Office of Scientific Research under award number FA2386-17-1-4660 and U.S. Army Ground Vehicle Systems Center. \end{funding}

\cleardoublepage

\setlength{\cftbeforetoctitleskip}{3.0em}
\setlength{\cftaftertoctitleskip}{2.0em}
\setlength{\cftbeforeloftitleskip}{3.0em}  
\setlength{\cftbeforelottitleskip}{3.0em}  
\cftsetpnumwidth{1.75em}  

\hypersetup{linkcolor=black}  
\hypertarget{contents}{}
\microtypesetup{protrusion=false}
\tableofcontents
\clearpage
\listoffigures
\clearpage
\listoftables
\microtypesetup{protrusion=true}

\mainmatter
\pagestyle{thesis}

\hypersetup{linkcolor=documentLinkColor}  


\onehalfspace

\chapter{Introduction} \label{secIntro}

\section{Background and Motivation}

When an intelligent agent works closely with human beings (e.g. robot navigating in a crowd), or co-exists with other mobile agents which are controlled by humans in an environment (e.g. autonomous vehicle interacting with other manned vehicles in traffic), it is critical for it to understand and predict the motions of other agents in the same environment to ensure safe interaction and efficient performance. 

In this work, we focus on the problem of pedestrian trajectory prediction in crowded environments. 

Humans naturally have the inherent ability to safely navigate in dense crowds, by recognizing other pedestrians' intentions, understanding the commonly accepted social norms, and reasonably predicting others' future motions. However, these instincts are not easily obtainable by robots or other intelligent agents. Predicting human pedestrians' future trajectories is actually a rather challenging task, because the future trajectories can be affected by not only the physical properties that we have well-established models to explain, such as energy or momentum, but also the pedestrians' hidden objectives and subtle social norms in crowd interactions.

\begin{figure}[ht]\centering
\includegraphics[width=4.7in]{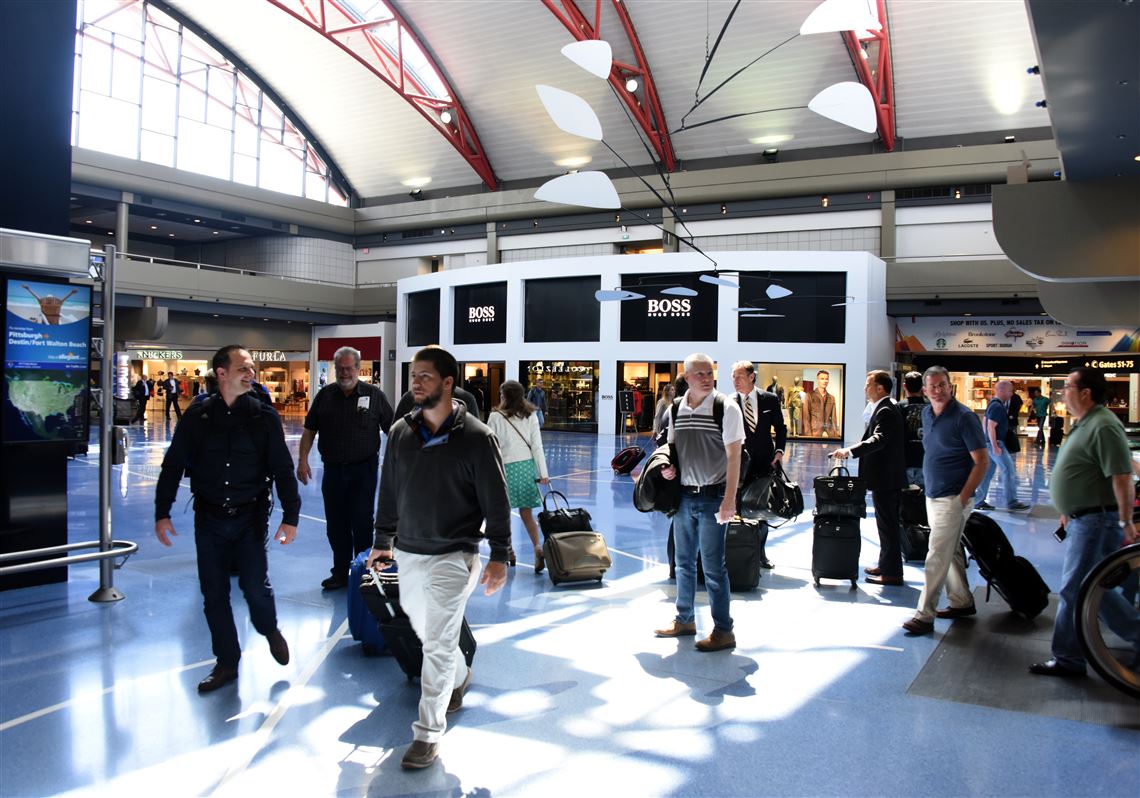} 
\caption{Pittsburgh International Airport (PIT)~\cite{PIT-airport-pg}. An example showing human motions in a typical crowd.}
\label{fig:conv_compare}
\end{figure}

\section{Motion Patterns}
Motion Patterns here are defined as short segments of pedestrian trajectories that can be repetitively observed in real recorded data.  

Our approach relies on two fundamental beliefs that we have: 
\begin{itemize}
    \item People's motions are affected by other people's motions, \figref{fig:4scenario}
    \item When other people's motions are seemingly different, if they fall under the same pattern, people often still react with the same future motions
\end{itemize}

The first statement is evidently true. We argue that the second statement is also true in most cases. For example, a person on your front left walking towards you at a very fast speed would likely make you want to slow down, slightly move to your right and give way to them. And, you would mostly likely do the same thing, no matter whether they are a few centimeters to their left or to their right, whether they are walking at 4.9 meters per second or 5.1 meters per second. Here, walking fast towards you from your front left is a motion pattern, and there can be many seemingly different but actually similar motions belonging to this pattern. Yet another way to see this is that, in the examples given in \figref{fig:4scenario}, others pedestrians' trajectories can vary by some reasonable amount, and that should not change our prediction about the green target agent's future motion very much. 

Supposing that an intelligent agent knows the set of motion patterns, and knows what people typically do to react to these patterns, at the time of predicting pedestrians' future trajectories, it can match the observed trajectories to the known patterns, and then accordingly gives the stored typical behaviors as predictions. This is the general idea that we are considering to solve the problem of human trajectory prediction.

\section{Social Pattern Extraction Convolution (Social-PEC)}

The majority of state-of-the-art works in pedestrian trajectory prediction generally are deep learning models nowadays, and they mostly follow the encoder-decoder model. In these models, history trajectories, also called observed trajectories, are passed into encoder modules, which typically are neural networks, e.g. Recurrent Neural Network (RNN). As a result, the information from these history trajectories will be represented in some latent space which unfortunately is not well understood most of the time. In the end, the prediction of future trajectories would be obtained by decoding information in latent space with another neural network module. 

Our approach also roughly follows the encoder-decoder model, while it avoids some questionable practices commonly found in existing works, and provides a more transparent and explainable decision making process at the same time.

We propose Social Pattern Extraction Convolution (Social-PEC). 

The model is designed to learn, recognize and utilize various motion patterns of walking pedestrians in dense crowds. It is forced to ``notice'' the motion patterns and predict upon them during training and inference. 

We build our sequence encoder using the idea of Temporal Convolutional Neural Networks (CNN)~\cite{lecun89cnn} and propose a new convolution operator (defined in \ref{sec:pattern})
that enables our model to actually detect, learn, and extract motion patterns from the observed trajectories. Using a different convolution operator is not a new idea: in~\cite{ghiasi19-generalizing}, the conventional correlation-based convolution operator has been modified to successfully achieve satisfying performance on the MNIST dataset, showcasing the applicability of the generalization of convolution operation in CNNs. 

Our model, Social-PEC, achieves comparable results with the state-of-the-art methods on public datasets in terms of standard evaluation metrics based on the displacement errors. Additionally, the use of motion patterns unveils the obscurity in social pooling, and makes the decision making process more transparent, intuitive, and explainable.


    \newcommand{\scaleA}{0.45}
    \begin{figure}[t]\centering
    \subfloat[]{\includegraphics[trim=0 0 732 0,clip,scale=\scaleA]{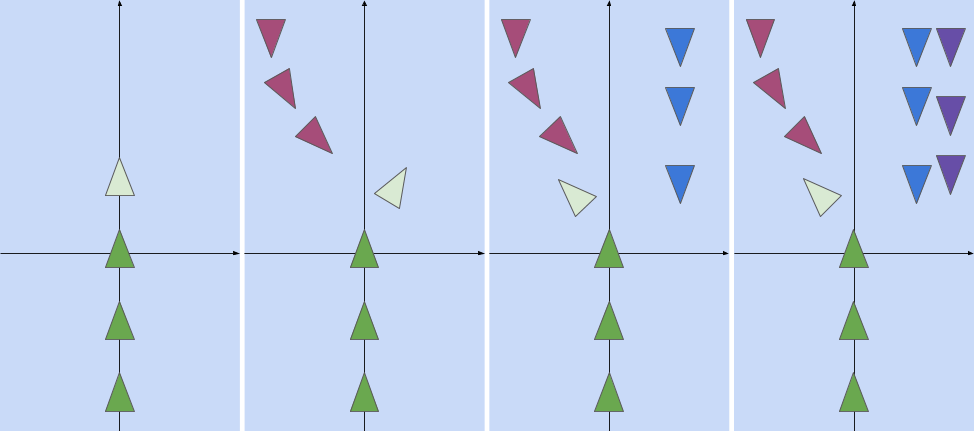}} 
    \subfloat[]{\includegraphics[trim=244 0 488 0,clip,scale=\scaleA]{figure/4scenario.png}}
    \subfloat[]{\includegraphics[trim=488 0 244 0,clip,scale=\scaleA]{figure/4scenario.png}}
    \subfloat[]{\includegraphics[trim=732 0 0 0,clip,scale=\scaleA]{figure/4scenario.png}} 
    \caption{Four sample scenarios of pedestrian interaction. The green triangles and the light green triangle represent the target pedestrian's history trajectories and expected one-step future location; triangles in other colors, the history trajectories of other pedestrians; and the sharp tips, the trajectory directions. In (a), no motion patterns from others are present; therefore, the target is expected to proceed linearly. In (b), the motion pattern of ``somebody approaching me from my front left'' is present, which impacts our target to walk to the right in order to avoid a possible collision. In (c) and (d), another motion pattern of ``somebody on my right is about to pass by me'' is present; therefore, our target pedestrian walks towards left. }
    \label{fig:4scenario}
    \end{figure}



\section{Thesis Organization}

The remainder of the thesis is organized as follows: 

Chapter \ref{secReview} reviews the existing approaches in this field; 

Chapter \ref{secMethod} formally defines the problem, and explains the prediction algorithm, Social-PEC, in full details;

Chapter \ref{secEval} presents some quantitative and qualitative evidences, suggesting that the proposed algorithm is actually capable of achieving state-of-the-art performance while avoiding the obscurity and providing better explainability;

Chapter \ref{secFture} discusses possible future extensions of this work;

Chapter \ref{secConclu} concludes the thesis.


\chapter{Literature Review} \label{secReview}

Several technical approaches have been used to tackle the pedestrian trajectory prediction problem. In early works, algorithm designers tried to assert domain knowledge about social interactions in crowds to algorithms explicitly, e.g., Social Force~\cite{helbing95-socialForce} and Interactive Gaussian Process (IGP)~\cite{Trautman10-igp, Trautman13-multiGoal}. However, some researchers later suggested that hand-crafting models and rules have various limitations. These findings have led to those approaches that would allow the machines to learn directly from data~\cite{kitani12-af, alahi16-socialLstm, gupta18-socialGan}, resulting in significantly improved performances in general. We will mainly discuss the latter data-driven type of techniques in this section. 

\section{The Wide Use of RNN/LSTM}

For modeling sequential data, Recurrent Neural Network (RNN) and its variations such as Long Short-Term Memory network (LSTM) have been the popular choice, e.g.,  Social-LSTM~\cite{alahi16-socialLstm}. However, the benefit or necessity of using RNNs for pedestrian trajectory prediction in this problem domain of predicting human future trajectory is debatable. 

RNNs tend to gradually forget information from the past, hence the idea of LSTM was proposed as a remedy to selectively forget/remember. An LSTM is indeed a reasonable choice in many problem domains, where the sequences can be arbitrarily and extensively long, e.g., in the text-related tasks, a sequence could be a lengthy article where important contextual information can appear anywhere in the text. 

By contrast, in the pedestrian trajectory prediction problem, except for some rare extreme cases, the models do not need excessively long sequences as their inputs, as it is generally enough to observe how people have moved in the last 5 seconds, i.e., the observations from too far past are no longer meaningful to the current interaction, if not misleading at times. 

Moreover, RNNs/LSTMs have other drawbacks~\cite{bai18-cnnrnn} including vanishing/exploding gradients, unstable and expensive training, and inefficient parameters. Remarkable efforts were also made by Bai et al.~\cite{bai18-cnnrnn} and Becker et al.~\cite{becker18-comparison} to empirically evaluate RNNs for sequential data learning.  

\section{Mysterious ``Social Pooling Layers''}

After each sequence is modeled and encoded, information needs to be aggregated together. 

Suppose that a scene has $N$ pedestrians, where $N$ is not a constant. In almost all of the existing approaches~\cite{alahi16-socialLstm,gupta18-socialGan,sadeghian19-sophie,kosaraju19-bigat}, for a model to predict one pedestrian's future trajectories, all other $(N-1)$ trajectories need to be gathered in a way that the dimension of $(N-1)$ collapses, i.e. a $(N-1)\times l$ tensor resulting in a length-$l$ vector, where $l$ is the length of encoded embeddings in latent space. For this purpose, the use of pooling layers has been a popular choice, where the $(N-1)\times l$ tensor becomes a length-$l$ embedding vector by maximum or average pooling along its second dimension.

Pooling layers are commonly found in Convolutional Neural Networks (CNN), especially with image-related tasks, typically after spatial convolutional layers~\cite{krizhevsky12-alexnet}. In these image CNNs, the pooling operation is clearly motivated and well understood: the purpose of pooling is to extract the strongest signal at a local region, and the vectors being ``pooled'' are the correlations between the kernels and inputs signals.

However, the latent space of RNNs/LSTMs is not as clear. It is rather difficult, if possible at all, to understand its semantic meanings in a physical space. Therefore, the use of pooling layers after RNN/LSTM's modelling sequences for information aggregation still lacks justifications. On this note, Mohamed et al.~\cite{mohamed20-stgcnn} reached consensus with us. 

\section{A reasonable choice: Graph Representation}

Another popular design choice is to use the graph representation, under which the information aggregation between multiple trajectories does not need to be done by pooling layers, which inherently avoids the obscurity of pooling in latent space. 

The combination of a graph with RNNs/LSTMs~\cite{vemula18-attention, kosaraju19-bigat}, and the combination of a graph with CNNs (Graph CNN)~\cite{mohamed20-stgcnn} have both been proposed. 

Theoretically it is a more sound solution to the problem in our opinion, and practically speaking, it did also achieve the best performance currently~\cite{mohamed20-stgcnn}. However, it is not the direction we decided to take. As we believe in graph representation's good potential, the size of a graph is generally dependent on the number of pedestrians in the scene, so it can face the scalability challenges as the number of pedestrians grows significantly in crowded scenes. Moreover, our approach offers not only theoretical soundness and state-of-the-art performance, but also a more transparent and explainable decision making process. 


\section{Overall Comments}

An extensive and comprehensive survey article for human motion prediction is done by Rudenko et al.~\cite{rudenko2019-predSurvey}, where interested readers can further find additional relevant works.

In contrast to the existing approaches, the proposed Social Pattern Extraction Convolution (Social-PEC) model avoids the issues mentioned above in this chapter. 

Firstly, it takes a fixed length of history trajectory as inputs and does not rely on RNNs to encode trajectory, thus avoids the RNN training issue. Secondly, the trajectory encoder is based on ``motion patterns'' that are intuitively reasonable and can be easily visualized, thus avoids the obscurity of ``social pooling'' when aggregating the encoded embedding of the neighboring pedestrians' trajectories. 
\chapter{Trajectory Prediction with PEC} \label{secMethod}

Our approach builds on an assumption that, when navigating in a crowd, humans react to an abstract representation of a scene, e.g., at the level of motion patterns that they have seen frequently in their past experiences. For instance, whether they are 2 or 3 people, whether they walk slightly faster or slower, whether they are a few centimeters to the left or right, as long as they approach from the same general direction at roughly the same distance with similar speeds, we probably will react very similarly, as shown in~\figref{fig:4scenario}. Although trajectory data are mostly stored as sequences of location coordinates, humans do not react to precise location coordinates; instead, we react to general \emph{motion patterns}. 

In this paper, we define a ``pattern'' to be a segment of data. Specifically, a ``motion pattern'' refers to a short segment of trajectory that can be frequently seen in real trajectory data. A motion pattern is represented as a sequence of location coordinates, similar to a short trajectory representation. 

Logistically, our strategy is to only predict one-step future locations, and use the predicted locations as if they are the new observations to further predict. When predicting a one-step future, we predict for each pedestrian at a time. 
The pedestrian we predict for is referred to as a ``target pedestrian'',
while all others are ``context pedestrians'', and all trajectories are transformed from the world coordinates to the target pedestrian's egocentric coordinates where the target pedestrian is at the origin, facing the positive direction of the x-axis.

In this section, we first clarify problem setup and notation; then introduce our model's key component, Pattern Extraction Convolution (PEC); next, we introduce the actual trajectory predictor; finally, we explain training and inference.

\section{Representation and Notation}\label{sec:problem}

Suppose that there are $M$ pedestrians in a scene. Given all of their observed history trajectories in the world coordinates, the goal is to predict the future trajectories of all pedestrians. 

A trajectory is a series of states timestamped at a constant interval. State $\sbf$ is defined as a 2-dimensional location coordinates, s.t., $\sbf=(x,y)$. We note that the definition of a state can be extended to include additional information such as  orientation or personality in the future.

In this paper, we use a finite length $T_f$ of timesteps. The start and end time for history observations are $1$ and $T_h$, and those of future prediction are $T_h+1$ and $T_f$.

For pedestrian $m \in \{1, ..., M\}$, the trajectory of $m$ observed from timestep $1$ to timestep $T_h$, denoted by $\phi^m_{1:T_h}$, is composed of a sequence of $(x,y)$ coordinates as follows: 
\begin{align}
    \phi^m_{1:T_h} &= \{\sbf^m_t|t\in\{1,..,T_h\}\}
\end{align} where $\sbf^m_t= (x^m_t,y^m_t)$ is pedestrian $m$'s state at time $t$.

Let $\Phi$ denote the set of trajectories, s.t., $\Phi[m]=\phi^m, \forall m \in \{1,...,M\}$; the dimension of $\Phi$ is thus $(M,T_f,2)$.

\[\Phi_{1:T_h} =\{\phi^m_{1:T_h} |m\in\{1,...,M\}\}\]
\[\hat\Phi_{T_h+1:T_f} =\{\hat\phi^m_{T_h+1:T_f} |m\in\{1,...,M\}\}\]
\[ \phi^m_{T_h+1:T_f} = \{\sbf^m_t|t\in\{T_h+1,...,T_f\}\}\]
\[\sbf_{T_h+1}^m\]
\[(T_f-T_h)\]

\section{Pattern Extration Convolution}\label{sec:pattern}

Pattern Extraction Convolution (PEC) is a mechanism that, in an intuitive sense, detects and recognizes patterns from data, while effectively projects trajectories from the x-y coordinate space to a new space that is defined in terms of similarities between patterns and the trajectory.

Let $\Pb$ denote the set of motion patterns, and let there be $N$ patterns in total. The $j$-th motion pattern, $\forall j\in\{1,...,N\}$, is defined as the following:
\begin{align}
    \Pb[j] &= \{\sbf^j_t|t\in\{1,..,L\}\}
\end{align}
where\\
$L$ is the pattern length,\\
$\sbf^j_t=(x^j_t,y^j_t)$ is the state. 

The dimension of $\Pb$ is thus $(N,L,2)$. 

We define PEC as an encoder that translates raw trajectory $\phi$ to abstract trajectory $\psi$ in terms of motion patterns $\Pb$ as follows: 
\begin{equation}\psi = \text{PEC}(\phi;\Pb)\end{equation}
At each timestep $t$, for each motion pattern $j$, the PEC operation is defined as:
\begin{align}\label{eq:pec}
    \psi[t,j] &= \text{PEC}(\phi;\Pb)[t,j] \\
        &=\lmdb[j]\cdot\log( \sum_k^L \Big\|\phi[t-L+k,:]-\Pb[j,k,:]\Big\|_2 ) + \mathbf{b}[j] \nonumber
\end{align}
where\\
$\lmdb$ is a scaling coefficient, and 
$\mathbf{b}$, biases. 

\begin{figure}[ht]\centering
    \subfloat[]{\includegraphics[width=2.8in]{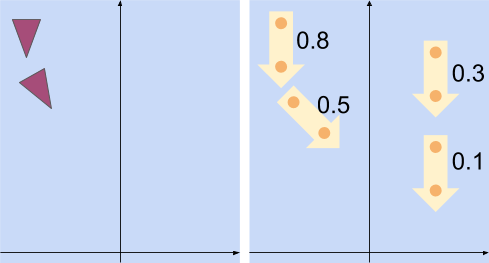}}\, 
    \subfloat[]{\includegraphics[width=2.8in]{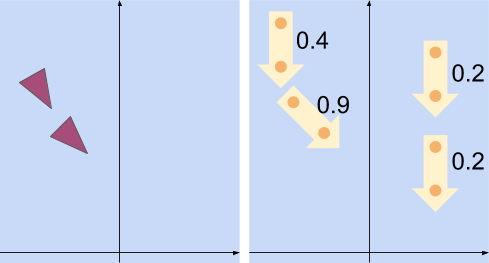}}\\ 
    \centering\subfloat[]{\centering\includegraphics[width=3in]{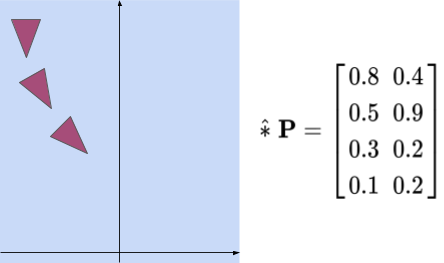}} 
    \caption{An example showing how raw trajectories can be projected to the motion pattern space. 
    Red triangles represent the input trajectory. The full trajectory $\phi$ is shown in (c) which is of shape $(3,2)$, for $T=3$. 
    Yellow arrows represent the set of motion patterns of shape $(4,2,2)$, for the number of patterns $N=4$, the length of pattern $L=2$, and $\sbf=(x,y)$ is 2-dimensional. 
    In (a), the similarities between the first segment of the trajectory and each motion pattern are found and marked next to the arrow; in (b), the second segment, similarly.
    The full operation on the trajectory level is shown in (c), where the resulting matrix is $\psi$ in the shape of (2,4) for $T_h-L+1=2$ and $N=4$. 
    The columns of this matrix $\phi$ are similarity scores from (a) and (b), respectively.%
    }
    \label{fig:conv}
\end{figure}

The dimension of the resulting encoded trajectory $\psi$ is $(T_h-L+1,N)$ where each entry of $\phi[t,j]$ indicates the similarity between the corresponding segment of the trajectory and the pattern $P[j]$. This operation is further demonstrated in \figref{fig:conv} with an example. 

Because the output $\psi$ of the PEC operation will later interact with an activation function such as \emph{tanh}, it is necessary to bring in the $\log$ function and the extra scaling coefficients $\lmdb$. Specifically, the $\log$ function helps re-range the values of the L2-distances from $[0,+\infty)$ back to $(-\infty,+\infty)$. Scaling coefficients $\lmdb$ helps scale the output properly to better interact with the non-linearity of the activation function. These two practices were not necessary for the conventional convolution operator
because the dot-product operation naturally ranges $(-\infty,+\infty)$ where the magnitude of a kernel matrix adds already an extra degree of freedom to help scale the response.


It is worth noting that the PEC operator defined in~\eqref{eq:pec} is different from the conventional convolution (CONV) operator~\cite{ghiasi19-generalizing,paszke19-pytorch} which is commonly used in CNNs for image-related tasks. The main differences are (1) PEC is based on the L2-difference to measure the physical distance whereas CONV is based on dot-product; (2) the conventional operator ignores the physical meaning of the channels and simply sums up the outputs from different channels. The necessity for using the PEC operator for trajectory encoding is further illustrated in~\figref{fig:conv_compare}.
%

The set of motion patterns, $\Pb$, is learned from data, trained using the prediction loss through back propagation. More implementation details and illustration are presented in Sections~\ref{sec:implementation} and~\ref{sec:show_steps}.

\newpage
\newpage

\begin{figure}[ht]\centering\includegraphics[width=4in]{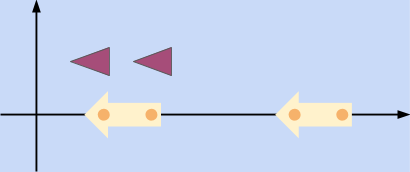} 
    \caption{An example showing the incompetency of the conventional convolutional operator for trajectories. Suppose red trajectory $\phi$ is \{(10,1),(20,1)\}, and two patterns $\p_0$ and $\p_1$ are \{(10,0),(20,0)\} and \{(50,0),(60,0)\}. Ideally, the magnitude of the convolution response (or output) is determined by the similarity only, i.e., the larger the more similar, but if CONV is used the magnitude of response would also be influenced by the magnitude of the input signals or kernels. In the example, the similarity between $\phi$ and $\p_0$ should be larger than $\p_1$, but $\text{CONV}(\phi,\p_0)=500<1700=\text{CONV}(\phi,\p_1)$.}
\label{fig:conv_compare}
\end{figure}

\section{Human Trajectory Predictor}\label{sec:model}

    \begin{algorithm}
        \newcommand{\Trans}{\texttt{Convert}}
        \newcommand{\TransBack}{\texttt{ConvertBack}}
        \newcommand{\LocPredictor}{\texttt{LocPredictor}}
        \begin{algorithmic}[1]
        \State $\Phi \gets \{\phi^m|m\in[1:M]\}$
        \Function{TrajPredictor}{$\Phi$}
            \For{$t=T_h+1,..,T_f$}
                \For{$m=1,..,M$}
                    \State $\Phi'\gets\Trans(\Phi[:,t-T_h:t], m)$
                    \State $\mu',\Sigma'\gets\LocPredictor(\Phi'[m],\Phi'[-m])$
                    \State $\sbf'\sim\mathcal{N}(\mu',\Sigma')$
                    \State $\sbf^m\gets\TransBack(\sbf') $
                    \State $\Phi[m,t]\gets\sbf^m$
                \EndFor
            \EndFor
        \State \Return $\Phi[:,T_h+1:T_f]$
        \EndFunction
        \end{algorithmic}
    \caption[.]{Trajectory Predictor}
    \label{algo:workflow}
    \end{algorithm}

    Our strategy is to predict one-step future locations first, and then use the predicted locations as if they are new observations to further predict. At each timestep, every pedestrian is treated as the target pedestrian by turns. 
    
    For each target pedestrian $m \in \{1, ..., M\}$, the location coordinates of all trajectories are first transformed from the world coordinates to the new coordinates, where pedestrian $m$ is at the origin, oriented towards the positive direction of the x-axis, as in the \texttt{Convert} function in~\algoref{algo:workflow}. 
    
    Next, the Location Predictor takes all of these trajectories as inputs, and outputs $m$'s next location, which is modeled as a bivariate Gaussian distribution~\cite{alahi16-socialLstm}. 
    
    Finally, this location is converted back to the world coordinates and appended to $\Phi$ for location prediction for the next timestep.
    
    \begin{figure}\centering
        \includegraphics[width=\textwidth]{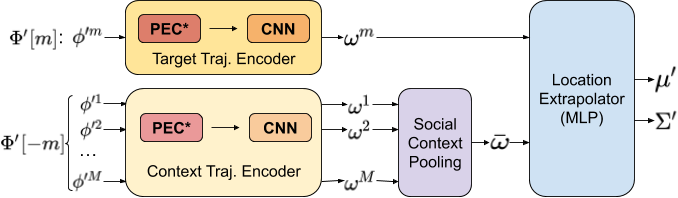}
        \caption{Location Predictor that predicts one-step location for pedestrian $m$. $\Phi'$ is the set of trajectories already transformed to target pedestrian's coordinates. $\omega$ is the trajectory embedding, explained in Equation~\eqref{eq:encoder}. }
        \label{fig:loc_predictor}
    \end{figure}

    Given target pedestrian $m$, let $-m$ denote the rest of the pedestrians except $m$, referred to as Context.
    First, $\Phi'[m]$ and $\Phi'[-m]$ are encoded respectively by two different encoder networks. The reason for having two different networks is that motion pattern set and context trajectories are expected to be different for Context and Target. The Trajectory Encoder works on one trajectory at a time. The Target trajectory embedding $\omega$ can be written as follows:
    \begin{align}\label{eq:encoder}
        \omega &= \text{CNN}(\sigma(\psi)) \nonumber\\
        &= \text{CNN}(\sigma(\text{PEC}(\phi;\Pb)))
    \end{align}
    where \\
    $\phi$ is the raw trajectory in location coordinates space;\\
    $\psi$, the trajectory encoded by PEC; and\\
    $\sigma$, the activation function. 
    
    By applying CNN to the encoded trajectory $\psi$, combinations of basic motion patterns are further extracted by the CNN on higher levels as more sophisticated patterns.
    
    Now that the model is ready to predict future trajectory for the target pedestrian, observations of all other pedestrians should be aggregated to provide the social context. The Context Pooling layer is applied to compute the context trajectory $\bar\omega$ as follows:
    \begin{equation}\label{eq:pool}
        \bar\omega[t,k] = \max_{m\in-m} (\omega^m[t,k])
    \end{equation}
    where \\
    $\bar\omega$ has the same shape as the target's encoded trajectory $\omega^m$, \\
    $t$ and $k$ are time and pattern indices. 
    
    To make predictions, both encoded target trajectory $\omega^m$ and social context $\bar\omega$ are fed into  Multilayer Perceptron (MLP) in the end:
    \begin{equation}\label{eq:mlp}
        x,y,a,b,c = \text{MLP}(\omega^m,\bar\omega).
    \end{equation}
    The $\mu'$ and $\Sigma'$ of the location's Gaussian distribution are then constructed from the raw MLP outputs:
    \begin{equation}\label{eq:gaussian}
        \begin{matrix}
            \mu' = [x,y]\\[0.3em]
            \sigma'_{xx}, \sigma'_{yy} = \exp(a), \exp(b)\\[0.3em]
            \sigma'_{xy} = \sigma'_{xx}*\sigma'_{yy}*\tanh(c)\\[0.4em]
            \Sigma' = \begin{bmatrix}\sigma'_{xx}&\sigma'_{xy}\\\sigma'_{xy}&\sigma'_{yy}\end{bmatrix}
        \end{matrix}
    \end{equation}
    where superscript $'$ indicates that the variable is in the target-centered coordinates instead of the world coordinates.
    
    The construction in \eqref{eq:gaussian} is necessary because the value range of the raw MLP outputs, $[-\infty,\infty]$, does not satisfy the constraints of the covariance matrices. 



\section{Uncertainty Modeling}

The real world is full of stochasticity. In this problem of pedestrian trajectory prediction, there are at least the following uncertainties which are worth discussing:
\begin{itemize}
    \item The inputs to the model are observed history trajectories of every pedestrian, which are typically obtained by some perception systems. These trajectories have been assumed to be reliable and accurate, but in reality, these data could be noisy due to the limitation of the perception system or natural physical constraints.
    \item The outputs of the model are supposed to match some ``ground truth'' future trajectories, as if these ground truth are the standard solution, but in reality, humans' future trajectories are multi-modal. That is, given the same set of history trajectories in a scene as inputs, there could be many sets of very different future trajectories that are all plausible and acceptable~\cite{gupta18-socialGan}.
    \item Uncertainty can also be caused by the limitations of the prediction model itself, e.g. the design of the model, the parameters of the model, or insufficient training, etc. This is also known as \textit{Epistemic Uncertainty} in the machine learning community~\cite{kendall17-bdl}. 
\end{itemize}

Here in this work, we attempt to model the second uncertainty mentioned above, by modeling the predicted future location with Gaussian Mixture Model (GMM), instead of one single Gaussian Distribution. 

In \figref{fig:loc_predictor}, as explained in \eqref{eq:mlp} and \eqref{eq:gaussian}, the output of the Location Predictor was a length-5 vector, representing a Bivariate Gaussian Distribution. If Gaussian Mixture Model is used to model the future location, the length of the location predictor output would then be $(5K-1)$, where $K$ is the number of mixture components. Under this assumption, the output should instead be:

\begin{align}   \label{eq:gmm}
    [&x_1,y_1,a_1,b_1,c_1,\\\nonumber
    &x_2,y_2,a_2,b_2,c_2,\beta_2\\\nonumber
    \dots\\\nonumber
    &x_K,y_K,a_K,b_K,c_K,\beta_K] = \text{MLP}(\omega^m,\bar\omega)
\end{align}

where $\beta_i$ is the weight of the $i$-th mixture component. Note that there is not $\beta_1$, because $\beta_1$ is assumed to be a constant, e.g. 1. Removing this redundant dimension is to ensure the stability of all the $beta$ values during training.

Mean values and covariance matrices, $\mu_i$ and $\Sigma_i$, for each mixture component are obtained in the same way as in \eqref{eq:gaussian}. 

The sampling procedure and the probability density function follow the typical operations of Gaussian Mixture Model. 

That still leaves the uncertainties caused by noise in the observed history trajectories and the limitation of the model unaddressed here. Our comments about these uncertainties can be found in the chapter \ref{secFture}.

\section{Training and Inference}

During training, the future prediction length is set to be 1, s.t., $T_f=T_h+1$,
because all parameters that need to be trained are all within the Location Predictor. Essentially, the whole model is trained as if it is a one-step future location predictor instead of a trajectory predictor. 
    
    Model parameters in the Location Predictor are learned by minimizing the negative log-likelihood loss shown below:
    \begin{equation}
        L = - \sum_{m=1}^M\log(\,P(\Phi'[m,T_h+1]\,\big\vert\,\mu'^m,\Sigma'^m)\,)
    \end{equation}

    During inference, the predictions of future trajectories can have an arbitrary length, and the prediction output is $\Phi[:,T_h+1:T_f]$ as stated in~\algoref{algo:workflow}. 

\chapter{Results} \label{secEval}

\newcommand{\bb}{\textbf}
    
\begin{table*}\centering
    \caption{Average Displacement Error (ADE) in meters for different methods, the lower the better. Linear Extrapolation is included as baseline and its output is deterministic, whereas all other models predict 20 trajectories at once, and only the best one is counted for evaluation.}
    \resizebox{0.9\textwidth}{!}{
    \begin{tabular}{c||c|c|c|c|c||c}
         Model        & ETH           & Hotel         & Univ.         & Zara1         & Zara2         & Ave.          \\\hline\hline
         
         Linear       
         & 1.33          & 0.39          & 0.82          & 0.62          & 0.77          & 0.79          \\\hline
         S-LSTM\cite{alahi16-socialLstm}       
         & 1.09          & 0.79          & 0.67          & 0.47          & 0.56          & 0.72          \\\hline
         SGAN(20VP20)\cite{gupta18-socialGan} 
         & 0.87          & 0.67          & 0.76          & 0.35          & 0.42          & 0.61          \\\hline
         STSGN\cite{zhang19-stsgn}        
         & 0.75          & 0.63          & 0.48          & \textbf{0.30} & \textbf{0.26} & 0.48          \\\hline
         S-BiGAT\cite{kosaraju19-bigat}      
         & 0.69          & 0.49          & 0.55          & \textbf{0.30} & 0.36          & 0.48          \\\hline
         S-STGCNN\cite{mohamed20-stgcnn}     
         & 0.64          & 0.49          & 0.44          & 0.34          & 0.30          & 0.44          \\\hline\hline
         \bb{Social-PEC}   
         & \textbf{0.61} & \textbf{0.31} & 0.47          & 0.43          & 0.35          & 0.43          \\\hline
         \bb{Social-PEC-GMM}   
         & 0.63          & 0.32          & \bb{0.42}     & 0.39          & 0.28          & \textbf{0.41} \\\hline
    \end{tabular}
    }
    \label{table:ade}
\end{table*}

\begin{table*}\centering
    \caption{Final Displacement Error (ADE) in meters for different methods, the lower the better. Linear Extrapolation is included as baseline and its output is deterministic, whereas all other models predict 20 trajectories at once, and only the best one is counted for evaluation.}
    \resizebox{0.9\textwidth}{!}{
    \begin{tabular}{c||c|c|c|c|c||c}
         Model        & ETH           & Hotel         & Univ.         & Zara1         & Zara2         & Ave.          \\\hline\hline
         
         Linear       
         & 2.94          & 0.72          & 1.59          & 1.21          & 1.48          & 1.59          \\\hline
         S-LSTM\cite{alahi16-socialLstm}       
         & 2.35          & 1.76          & 1.40          & 1.00          & 1.17          & 1.54          \\\hline
         SGAN(20VP20)\cite{gupta18-socialGan} 
         & 1.62          & 1.37          & 1.52          & 0.68          & 0.84          & 1.21          \\\hline
         STSGN\cite{zhang19-stsgn}         
         & 1.63          & 1.01          & 1.08          & 0.65          & 0.57          & 0.99          \\\hline
         S-BiGAT\cite{kosaraju19-bigat}      
         & 1.29          & 1.11          & 1.32          & 0.62          & 0.75          & 1.00          \\\hline
         S-STGCNN\cite{mohamed20-stgcnn}    
         & \textbf{1.11} & 0.85          & 0.79          & 0.53          & \textbf{0.48} & \textbf{0.75} \\\hline\hline
         \bb{Social-PEC}   
         & \textbf{1.11} & 0.52          & 0.82          & 0.77          & 0.60          & 0.76         \\\hline
         \bb{Social-PEC-GMM}   
         & 1.33          & \textbf{0.50} & \bb{0.75}     & 0.62          & 0.55          & \textbf{0.75} \\\hline
    \end{tabular}
    }
    \label{table:fde}
\end{table*}

\begin{table}[]\end{table}

\section{Datasets and Metrics}
Our model is evaluated on two datasets:~\cite{pellegrini09-eth} and~\cite{lerner07-ucy}. They contain 5 crowd sets in different scenes with a total number of 1,536 pedestrians exhibiting complex interactions such as walking together, groups crossing each other, joint collision avoidance and nonlinear trajectories. 

As for metrics, like~\cite{alahi16-socialLstm, gupta18-socialGan, zhang19-stsgn, kosaraju19-bigat, mohamed20-stgcnn}, we use Average/Final Displacement Error (ADE/FDE)~\cite{mohamed20-stgcnn}, which have been conventionally used for this problem in this research community. 

Average Displacement Error (ADE) is the mean square error (MSE) over all estimated locations of a trajectory and the ground truth:
\begin{align}
    \text{ADE} = \frac{\sum_{m=1}^M \sum_{t=T_h+1}^{T_f} \Vert\hat\sbf^m_t-\sbf^m_t\Vert_2 } {M(T_f-T_h)}
\end{align} where \\
$\sbf^m_t= (x^m_t,y^m_t)$ is pedestrian $m$'s state at time $t$, \\
$\hat\sbf$ and $\sbf$ represents prediction and ground truth respetively, \\
$T_h$ and $T_f$ are timestamps of the end of observation and the end of prediciton,\\
M is the number of pedestrians in the scene.

Final Displacement Error (FDE) is the distance between the predicted final destination and the ground truth final destination at the last timestep of the predicted future trajectory:

\begin{align}
    \text{FDE} = \frac{\sum_{m=1}^M \Vert\hat\sbf^m_t-\sbf^m_t\Vert_2 } {M(T_f-T_h)},\;t=T_f
\end{align}

In order to make full use of the data for evaluation and also to evaluate how well models generalize to unseen datasets, we use the leave-one-out approach where a model is trained and validated on 4 datasets and tested on the remaining set. To ensure a fair comparison, we use identical dataset step and train/validation/test split, which are also used in~S-SLSTM\cite{alahi16-socialLstm}, S-GAN\cite{gupta18-socialGan} and S-STGCNN\cite{mohamed20-stgcnn}.

The data used in our work are annotated every 0.4 seconds. Observation length is set to be 8 timesteps (3.2 sec) and future prediction length is set to be 12 timesteps (4.8 sec).

\section{Implementation Details}\label{sec:implementation}
As shown in \figref{fig:loc_predictor}, our model mainly contains 3 modules, Context Trajectory Encoder, Target Trajectory Encoder, and Location Extrapolator. 

The Context Trajectory Encoder consists of a Pattern Extraction Convolution (PEC) layer and a conventional convolution (CONV) layer, each followed by activation function \emph{tanh}, with a max pooling layer in between. For the PEC and CONV, the number of kernels are 100 and 160, the kernel lengths are 2 and 2; the pooling stride is 2. If the number of input channels is 2 (x-y coordinates) and the data temporal length is 8 (length of observations), s.t. an input is in the shape of (2,8), the resulting output's shape will be (160,3). 

The Target Trajectory Encoder is very similar to the Context Trajectory Encoder, except that the number of kernels are only 50 and 80, because there are many fewer varieties among target trajectories because the irrelevant variance has been removed by transforming the coordinate system with respect to the target trajectory in \texttt{Convert} of \algoref{algo:workflow}.

The Location Extrapolator are 4 fully-connected layers, followed by leaky Re-LU activation. The widths of the layers are 300, 120, 80, and 5. 

For training, the batch size is 64, and it trains for 150 epochs using the Adam Optimizer~\cite{kingma14-adam} with the learning rate set as 0.001. The model is trained on GeForce RTX 2080 Ti.

\section{Quantitative Results}
As shown in Table 1, our model outperforms almost all models and performs comparably well with the current state-of-the-art, Social-STGCNN~\cite{mohamed20-stgcnn}. Other works are mostly RNN-based, while ours and Social-STGCNN are CNN-based. 

Our results suggest that CNNs might be a better option for modeling some types of sequences, e.g., pedestrian trajectories; more in-depth discussions related to the comparison between CNN and RNN can also be found in~\cite{becker18-comparison} and~\cite{bai18-cnnrnn}. 


\section{Qualitative Analysis}

\begin{figure}[H]\centering
    \stackunder[0pt]{\includegraphics[trim=.56in .32in .63in .58in, clip,width=2.8in]{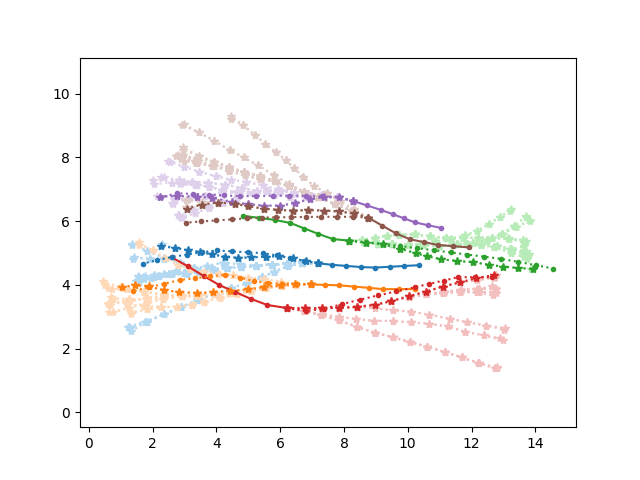}}{\footnotesize{(a)}}
    \stackunder[0pt]{\includegraphics[trim=.95in 0.52in 1.05in .95in, clip,width=2.8in]{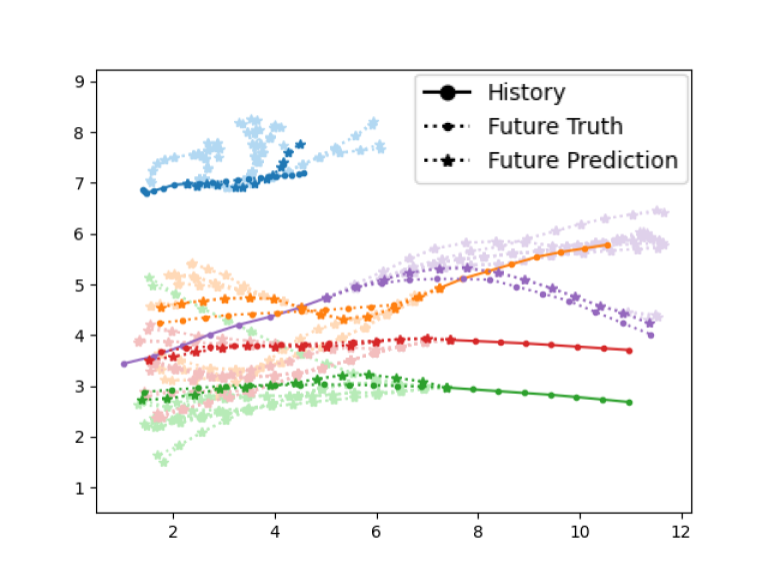}}{\footnotesize{(b)}}
    \stackunder[0pt]{\includegraphics[trim=.56in .32in .63in .58in, clip,width=2.8in]{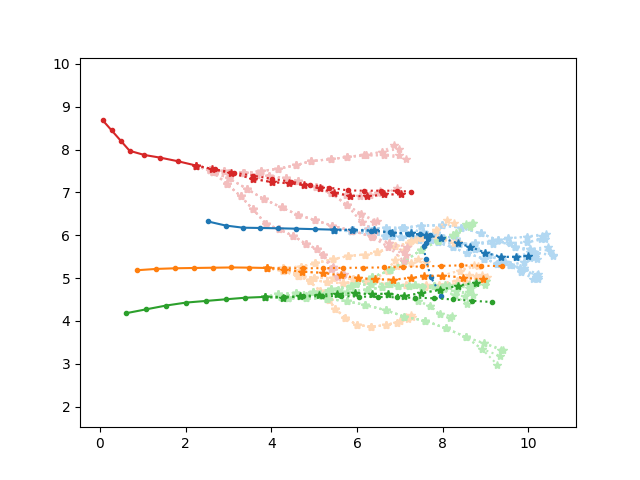}}{\footnotesize{(c)}}
    \stackunder[0pt]{\includegraphics[trim=.56in .32in .63in .58in, clip,width=2.8in]{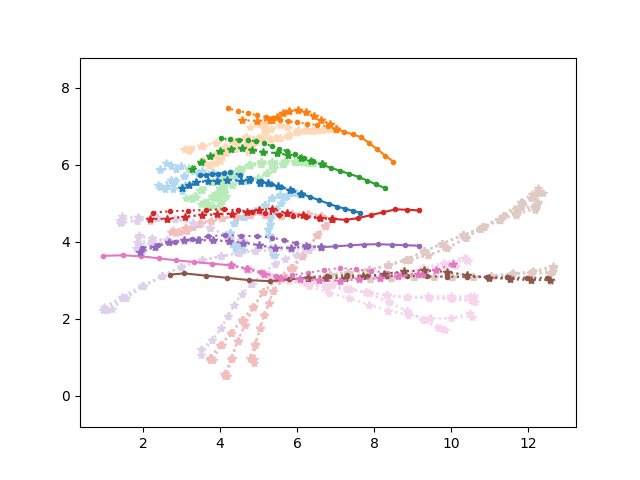}}{\footnotesize{(d)}}
    \stackunder[0pt]{\includegraphics[trim=.56in .32in .63in .58in, clip,width=2.8in]{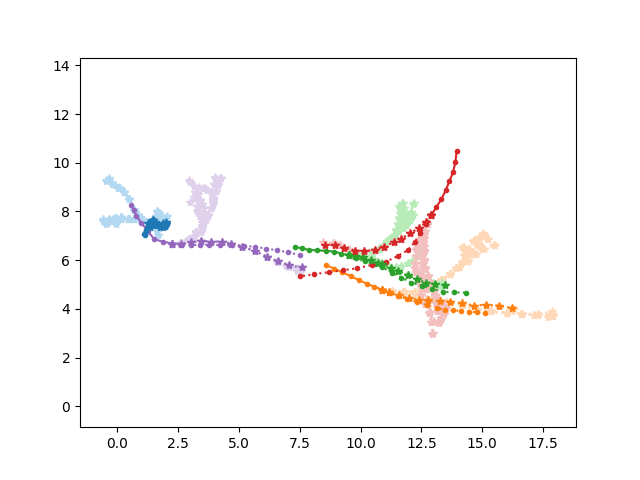}}{\footnotesize{(e)}}
    \stackunder[0pt]{\includegraphics[trim=.56in .32in .63in .58in, clip,width=2.8in]{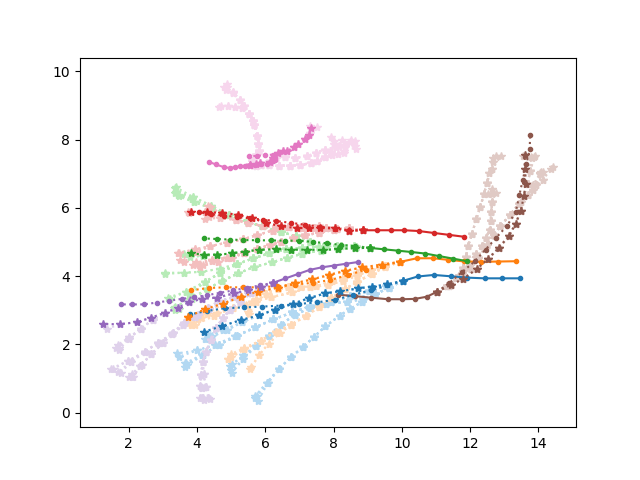}}{\footnotesize{(f)}}
    \caption{Some samples of our model's prediction. 
    Our model predicts 20 possible future trajectories for each pedestrian, of which the one with the smallest ADE is highlighted. This figure is best viewed in color. }
    \label{fig:results}
\end{figure}

Some sample results are shown in \figref{fig:results}. 
Typically, linear trajectories are trivial to predict; however, our results seem to support that the proposed model also performs well for some non-linear trajectories, especially in more crowded scenes, e.g., red in (a), purple and orange in (b), and brown in (f). 
The success here might indicate that the proposed model is able to make use of social context effectively to make more accurate and more reasonable predictions.

Some of the predictions deviate from the true future trajectories significantly, e.g., brown in (a), red in (e), and pink in (f). These predictions, however, still appear reasonable, that is, based on the observed trajectories, the prediction may appear arguably more reasonable than the true future. Such ``mispredictions'' are inevitable to some extent as some of the observed history trajectories might not carry enough information to allow anyone to accurately predict their future. 

Sometimes, the proposed is able to recognize the groups in a crowd and predict accordingly, although we did not explicitly design the model to incorporate social group awareness~\cite{yao19-group}. In (f), the history observations of orange and blue are highly similar, thus their future trajectories are predicted to be very similar too, even the ways how they deviate from ground truth are also similar. Comparatively, also in (f), the history observations of red and green did not demonstrate enough similarity, thus in the model's prediction they do not walk together any more. 

In some cases, the model successfully shows appropriate precaution for collision avoidance. In (e), the interaction between red and green is notable. Red took a different path that is farther away from green and both were predicted to move slower than ground truth. A plausible explanation is that the proposed model was trying to avoid the two pedestrians colliding with each other. In the same scene, the orange is predicted to move faster than ground truth, because its front space appears to be clear enough to allow faster speed.

\newpage

\newpage

\section{Motion Pattern Illustration}\label{sec:show_steps}

\begin{figure}[t]\centering
    \stackunder[0pt]{\includegraphics[trim=.76in .65in .76in .96in,clip,width=2.8in]{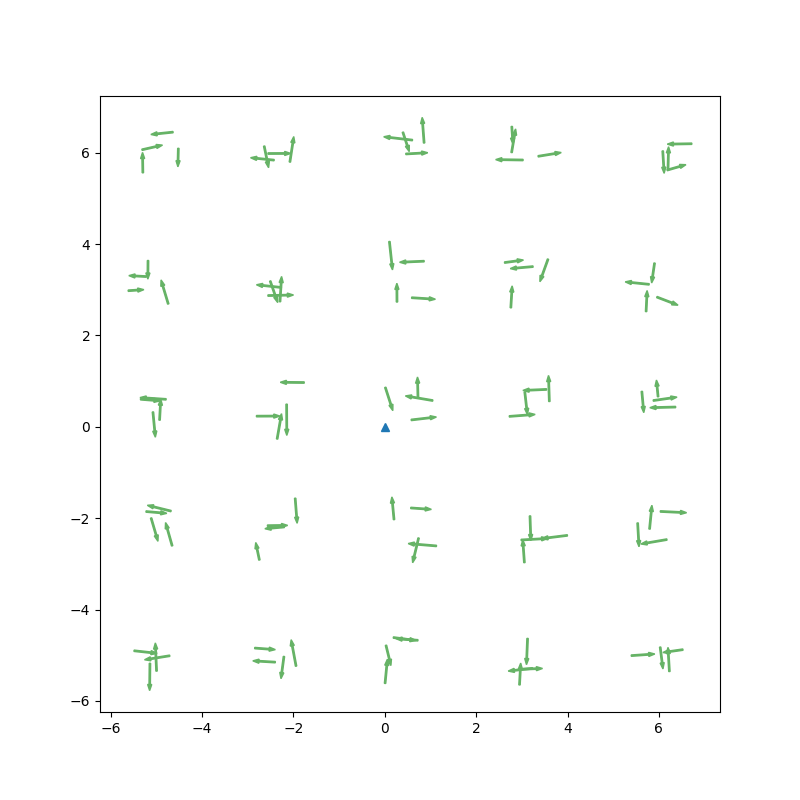}}{\footnotesize{(a)}}
    \stackunder[0pt]{\includegraphics[trim=.76in .65in .76in .96in,clip,width=2.8in]{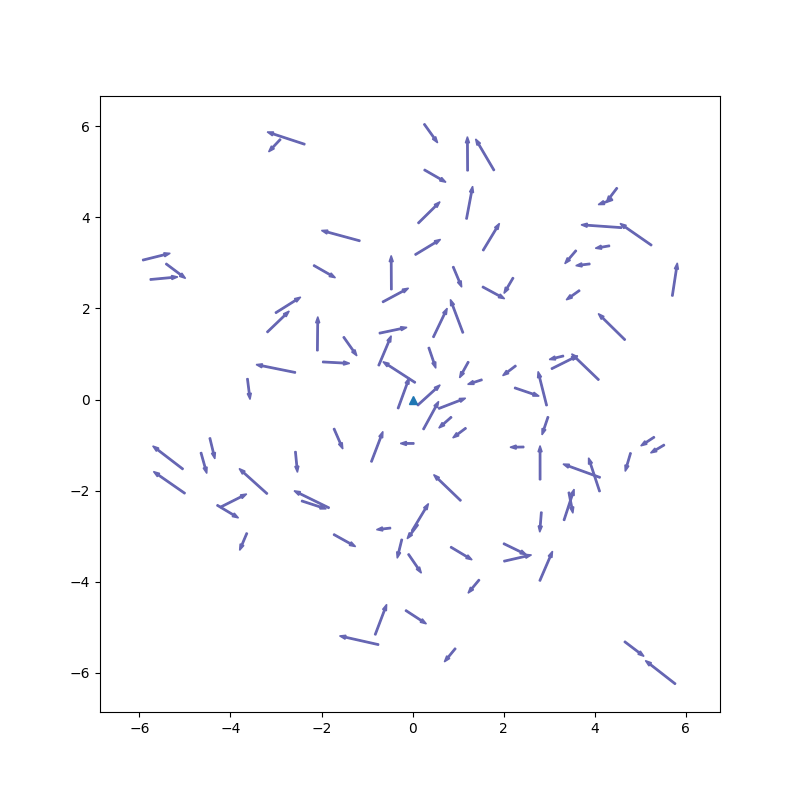}}{\footnotesize{(b)}}
    \stackunder[0pt]{\includegraphics[trim=.76in .65in .76in .96in,clip,width=2.8in]{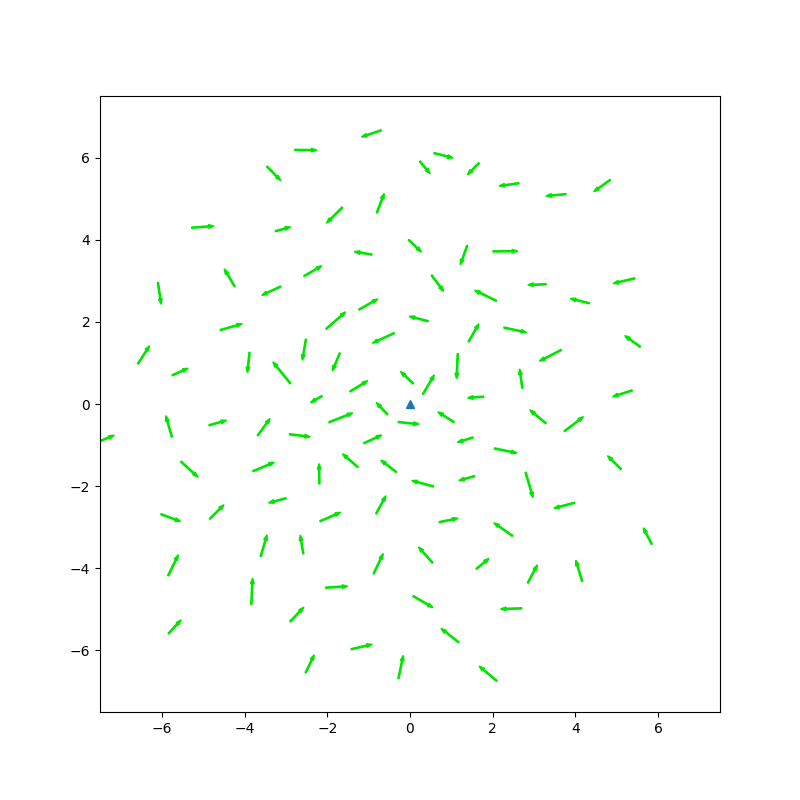}}{\footnotesize{(c)}}
    \stackunder[0pt]{\includegraphics[trim=.76in .65in .76in .96in,clip,width=2.8in]{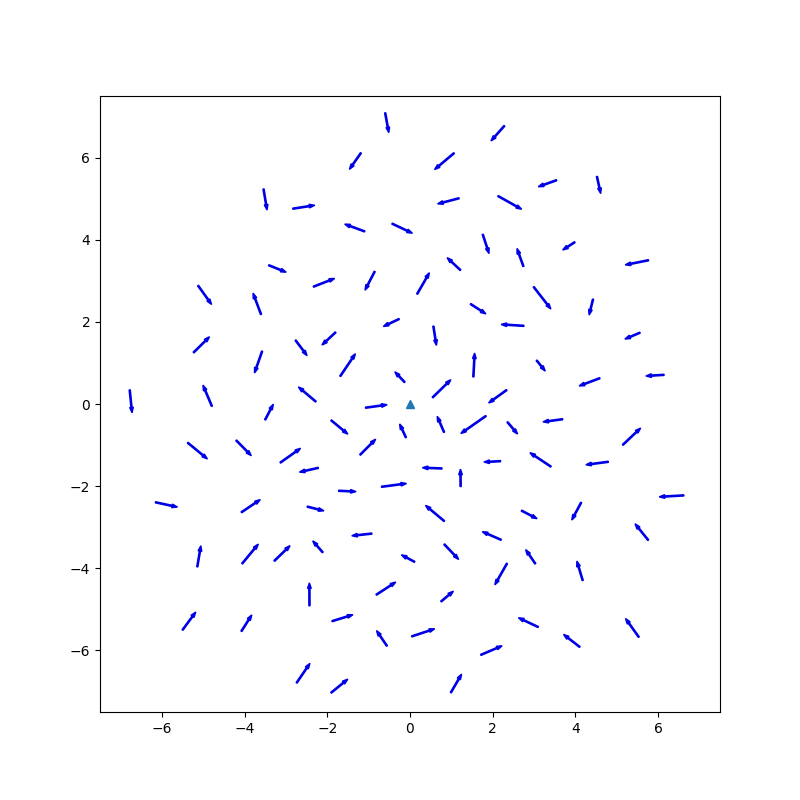}}{\footnotesize{(d)}}
    \caption{Two sets of initialization and learned motion patterns for the Context Trajectory Encoder. (a) and (b) are initialization, while (c) and (d) are their corresponding learned patterns after training. The number of patterns is 100; the length of patterns is 2. The coordinates are with respect to the target pedestrian marked by a blue triangle at (0,0). }
    \label{fig:init}
\end{figure}

\begin{figure*}[ht]\centering
    \includegraphics[width=6in]{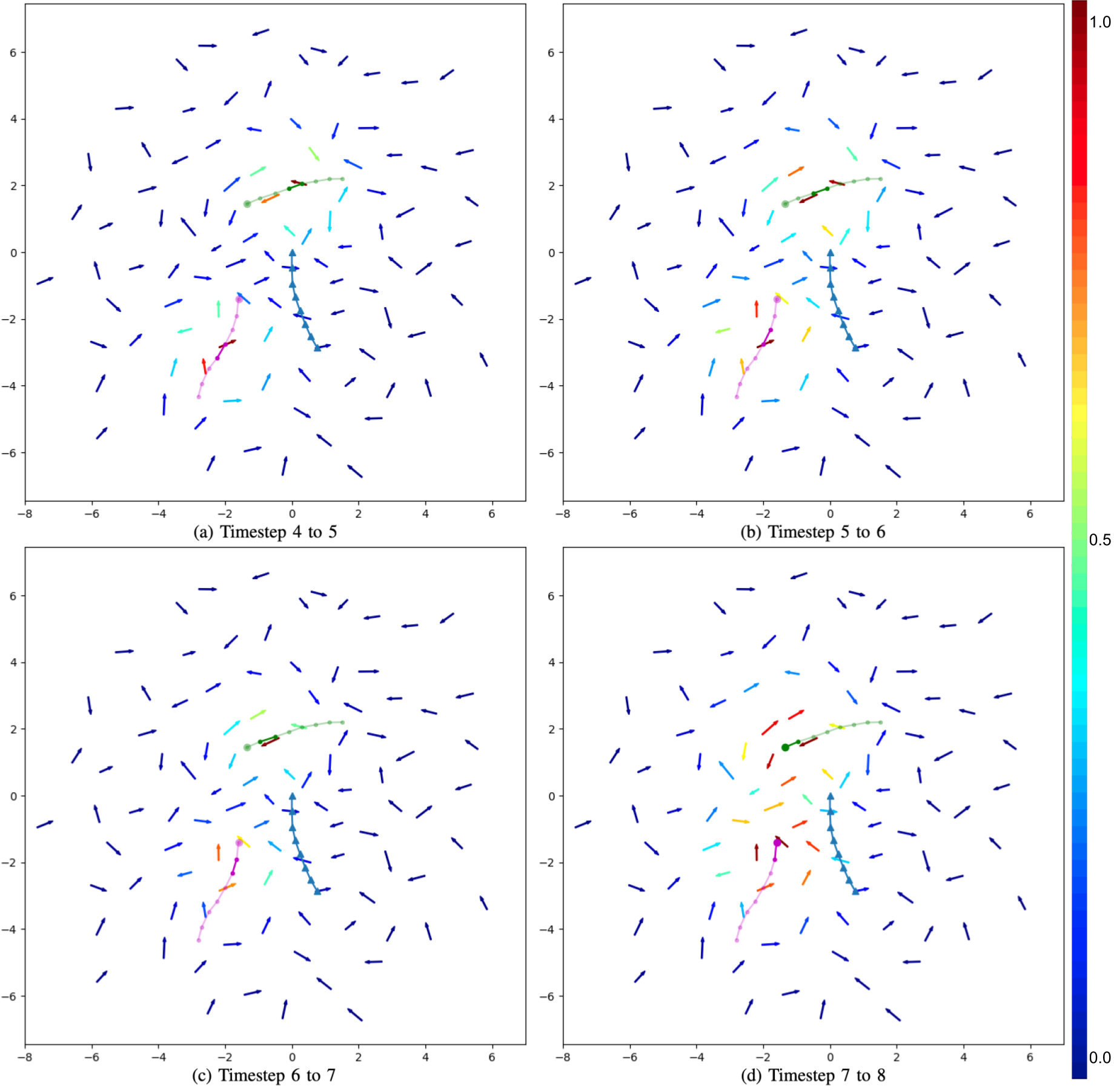}
    \caption{``Noticing motion patterns'' when presented with trajectory data. The arrows are from \figref{fig:init}(c). All trajectories are observed history. Dots are context trajectories and larger round dots indicate the last-timestep locations. Blue triangles are the target trajectory, which is only plotted but not matched to the learned patterns because patterns here are for context trajectories only. Trajectories are 8-timestep long, but the first 3 timesteps are neglected. Each subgraph demonstrates the matching of the particular trajectory segment, e.g., (d) shows the last segment, (c) shows the 2nd last, etc. By Pattern Extraction Convolution (PEC) layer, the segment is matched to similar patterns, causing these ``noticed patterns'' to be shown in heated color. }
    \label{fig:show_steps}
\end{figure*}

The weights for the PEC layer in the Trajectory Encoder are motion patterns. \figref{fig:init} presents two sets of patterns trained from two very different initializations, which demonstrates that the training is robust enough to allow different initializations to converge to similar patterns. It is noted that, though similar, the learned patterns are not identical; however, no significant performance difference has been observed in terms of ADE/FDE.

\figref{fig:show_steps} illustrates the core idea of ``Noticing Motion Patterns'' with an example scenario, where observed history trajectory data are matched to learned motion patterns at each timestep. This is done by the first layers of Trajectory Encoders, Pattern Extraction Convolution (PEC) layers. 

 Under the scheme of ``motion pattern'', social pooling finally is no longer obscure. We are not pooling in any unknown latent space, but pooling in well-understood pattern space. The physical meaning of each entry in tensor $\omega$ is the similarity indicating how much of that particular motion pattern is present in the current scene. The larger the entry value is, the more similar the raw trajectory is to the motion pattern. Thus, stronger motion patterns should have a bigger impact on the target pedestrian's decision making. By only considering the prominent presence for each motion pattern, the model can already be well-informed about its social context, of which \figref{fig:4scenario}(c)(d) are good examples.

\section{Applications}
Social robot navigation as a research field has been drawing attention from scientists and engineers for decades. A thorough survey on this topic can be found in~\cite{mavrogiannis21-core}. Understanding how human pedestrians interact with each other in crowds is core to this problem, where the presented work can be helpful. More concretely:
\begin{itemize}
    \item A crowd simulator is very important for evaluating and training social navigation algorithms. Human trajectory prediction algorithms, like Social-PEC, can be used to simulate pedestrians in those environments. In fact, the proposed Social-PEC has been adopted in the Social Navigation Simulator developed by Shum et al.~\cite{sns-repo}.
    \item With the appropriate constraints, Social-PEC has the potential to be extended or modified as a social navigation algorithm. Inspiration can be drawn from the previous work, NaviGAN by Tsai and Oh~\cite{tsai20}, in which the trajectory encoder and decoder can be replaced by the proposed Social-PEC, while safety and comfort can still be enforced by NaviGAN's social force module. 
\end{itemize}

\section{Embodiment: Social Navigation Robot ``Rocky''}
\begin{figure}[ht]\centering\includegraphics[width=\textwidth]{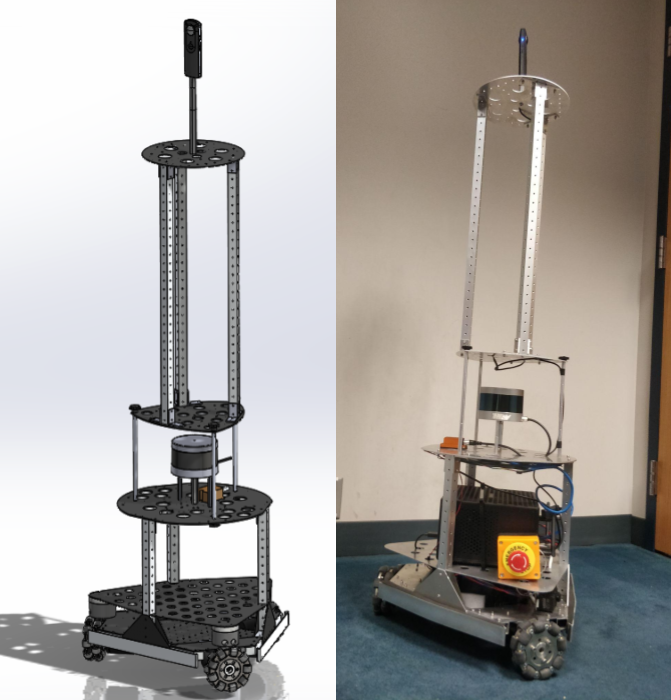} 
    \caption{The CAD model and picture of the social navigation robot ``Rocky''.}
\label{fig:rocky}
\end{figure}

In order to collect data and test the algorithm in the real world, we also designed and built a mobile robot platform, named ``Rocky'' \figref{fig:rocky}. It was built from scratch and it is fully functional now. 

    \subsection{Mobility}
    The robot is equipped with three brushless DC motors and three Omni-wheels, and has the max linear speed to any direction higher than $10$ meter/sec. A demonstration of its mobility is here: \url{https://youtu.be/N_oVchc29FQ}.
    
    \subsection{Perception}
\begin{figure}[t]\centering
    \includegraphics[scale=0.15]{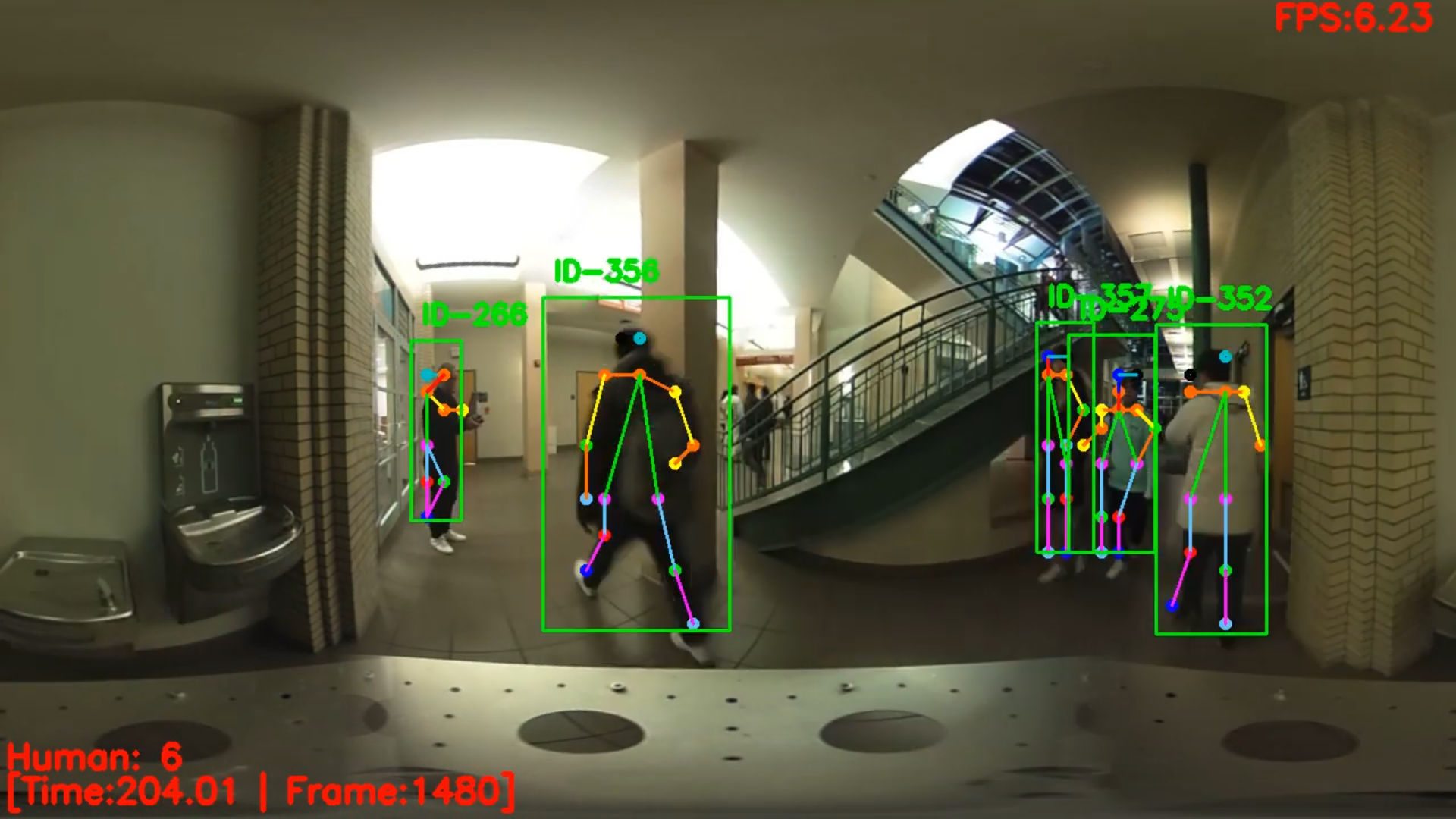}\\
    \includegraphics[scale=0.15]{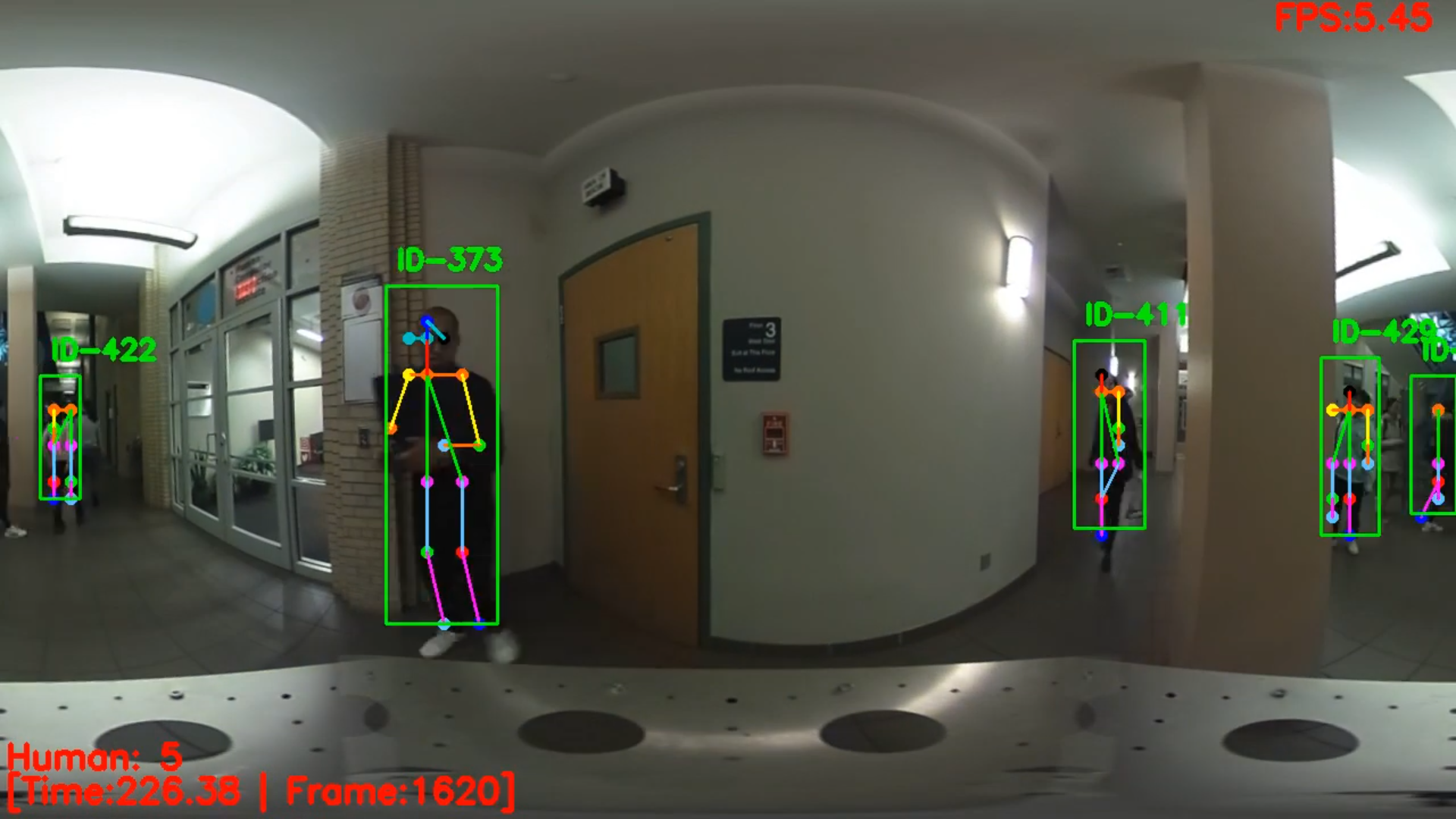}\\
    \includegraphics[scale=0.15]{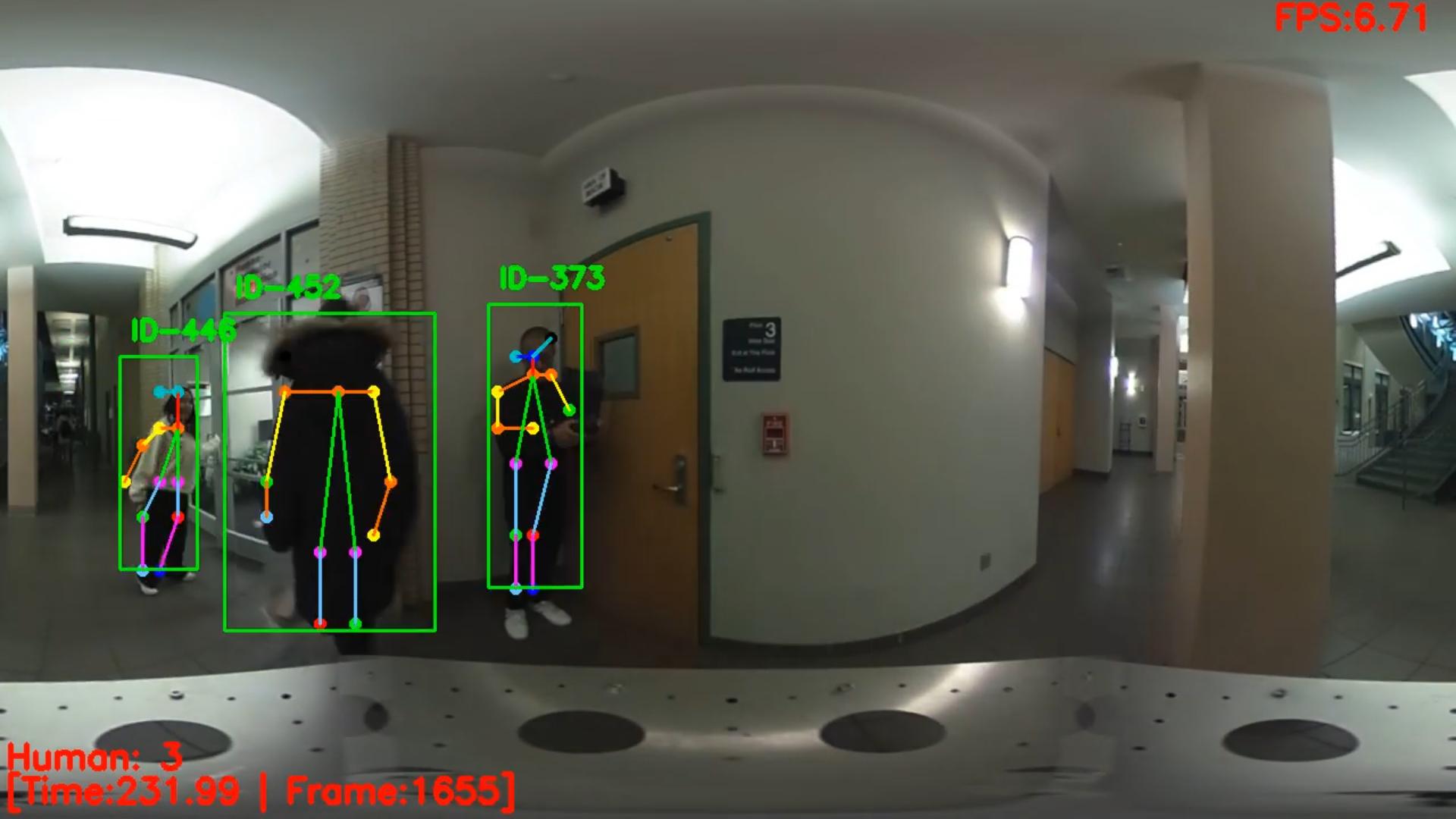}
    \caption{Real-time performance of the person tracking system. }
    \label{fig:tracking}
    \end{figure}

    The robot has a powerful sensing system:
    \begin{itemize}
        \item 1 3D LiDAR scanner, Velodyne VLP-16
        \item 1 IMU, Xsens MTi-30
        \item 1 360-degree dual-fisheye camera, Theta S
        \item 3 groups of microswitches connecting to bottom bumper
        \item 3 rotatory encoders on motors
    \end{itemize}
    It is capable of mapping the environment, detecting and tracking human pedestrians~\figref{fig:tracking}, state estimation, etc. The person tracking system used the open sourced OpenPose~\cite{cao19openpose}. A demonstration of its person tracking performance is here: \url{https://youtu.be/o-vwtU1uYeE}. 

\chapter{Future Directions} \label{secFture}

On this particular problem domain that we are working on, we believe our work can be further improved in the following ways:
\begin{itemize}
    \item No longer assume the inputs to the model are accurate and reliable, instead assume there is noise in the observations of history trajectories. Noise can be simply modeled as Gaussian noise or modeled by an additional neural network. For evaluation, noisy data can be simulated by adding randomly generated noise, or some actually noisy data. 
    \item Frame this problem as Bayesian Deep Learning~\cite{kendall17-bdl}, which inherently addresses and reports uncertainties of the input data and of the model. 
    \item All trajectories/locations are currently represented in the Cartesian coordinate system, $(x,y)$, while the polar coordinate system, $(\theta,r)$, potentially can be a better choice, because in the latter, $\theta$ and $r$ naturally decouple the distance and orientation of other pedestrians to the target pedestrian. 
    \item The idea of motion patterns should also be helpful in other types of motion than human motions, e.g. vehicles, aircrafts, etc. With the appropriate trajectory representation, the proposed idea might be able to be extended to other problem domains. 
    \item Currently, this work has assumed the environment to be open space with no physical obstacle, which is in fact not a realistic expectation. In the future, we can try to incorporate physical constraints of the environments into the human trajectory prediction model, which can potentially further improve its prediction performance. 
\end{itemize}


\chapter{Conclusions} \label{secConclu}

In this work, we propose a CNN-based model for human pedestrian trajectory prediction with the idea of motion patterns. 
The main contributions of this work:
\begin{itemize}
    \item we present Pattern Extraction Convolution (PEC), whch is used to encode trajectories in this work, as an intuitive and explainable mechanism to learn, detect, and extract patterns from data;
    \item we further apply PEC to the human trajectory prediction problem as the model of Social-PEC, and achieve comparable performance to the current state-of-the-art;
    \item the use of PEC avoids the obscurity in information aggregation (pooling layer) that was present in the previous work; and
    \item this study further challenges the community to re-examine the use of RNN in sequential data learning tasks.
\end{itemize}

\appendix

\chapter{Code and Publication} \label{secAppend}

\section{Code}
\url{https://github.com/cmubig/SPEC}

\section{Publication}
Dapeng Zhao and Jean Oh. Noticing motion patterns: Temporal CNN with a Novel Convolution Operator for Human Trajectory Prediction. \textit{IEEE Robotics and Automation Letters}, 2020. (\cite{zhao2020noticing})


\backmatter

\singlespace



\bibliographystyle{plainnat}

\bibliography{references}
\end{document}